\documentclass[lettersize,journal]{IEEEtran}
\usepackage{amsmath,amsfonts}
\usepackage{algorithmic}
\usepackage{array}
\usepackage[caption=false,font=normalsize,labelfont=sf,textfont=sf]{subfig}
\usepackage{textcomp}
\usepackage{stfloats}
\usepackage{url}
\usepackage{verbatim}
\usepackage{graphicx}
\usepackage{booktabs} 
\usepackage[table]{xcolor}
\usepackage{multirow}

\def\BibTeX{{\rm B\kern-.05em{\sc i\kern-.025em b}\kern-.08em
    T\kern-.1667em\lower.7ex\hbox{E}\kern-.125emX}}
\usepackage{balance}
\begin{document}
\title{GA-GS: Generation-Assisted Gaussian Splatting \\ for Static Scene Reconstruction}
\author{Yedong Shen$^{1 *}$, Shiqi Zhang$^{2 *}$, Sha Zhang$^{3, \dagger}$, Yifan Duan$^{1}$, Xinran Zhang$^{1}$, Wenhao Yu$^{4}$, \\ Lu Zhang$^{5}$, Jiajun Deng$^{6}$, Yanyong Zhang$^{2, \dagger}$~\IEEEmembership{Fellow,~IEEE}
\thanks{$*$ These authors contributed equally.}
\thanks{$\dagger$ Corresponding author}
\thanks{$^{1}$~School of Computer Science and Technology, University of Science and Technology of China, Hefei 230026, China {\tt\small \{sydong2002, dyf0202, zxrr\}@mail.ustc.edu.cn}}
\thanks{$^{2}$~School of Artificial  Intelligence and Data Science, University of Science and Technology of China, Hefei 230026, China {\tt\small zhangshiqi\_1127@mail.ustc.edu.cn, yanyongz@ustc.edu.cn}}
\thanks{$^{3}$ The Chinese University of Hong Kong, Hong Kong 999077, China,  {\tt\small zhangsha2048@gmail.com}}
\thanks{$^{4}$Institute of Advanced Technology, University of Science and Technology of China, Hefei 230026, China {\tt\small wenhaoyu@mail.ustc.edu.cn}}
\thanks{$^{5}$ Institute of Artificial Intelligence, Hefei Comprehensive National Science Center, Hefei 230088, China {\tt\small luzha@ustc.edu.cn}}
\thanks{$^{6}$ School of information Science and Technology, University of Science and Technology of China, Hefei 230026, China {\tt\small dengjj@ustc.edu.cn}}
}


\newcommand{\systemname}{GA-GS}
\newcommand{\truth}{authenticity} 
\newcommand{\ourdata}{Trajectory-Match Dataset}

\maketitle

\begin{abstract}
Reconstructing static 3D scene from monocular video with dynamic objects is important for numerous applications such as virtual reality and autonomous driving. 
Current approaches typically rely on background for static scene reconstruction, limiting the ability to recover regions occluded by dynamic objects.
In this paper, we propose \textbf{\systemname}, a \underline{\textbf{G}}eneration-\underline{\textbf{A}}ssisted \underline{\textbf{G}}aussian \underline{\textbf{S}}platting method for Static Scene Reconstruction. The key innovation of our work lies in leveraging generation to assist in reconstructing occluded regions. We employ a motion-aware module to segment and remove dynamic regions, and then use a diffusion model to inpaint the occluded areas, providing pseudo-ground-truth supervision. To balance contributions from real background and generated region, we introduce a learnable \truth\ scalar for each Gaussian primitive, which dynamically modulates opacity during splatting for \truth-aware rendering and supervision. Since no existing dataset provides ground-truth static scene of video with dynamic objects, we construct a dataset named Trajectory-Match, using a fixed-path robot to record each scene with/without dynamic objects, enabling quantitative evaluation in reconstruction of occluded regions. Extensive experiments on both the DAVIS and our dataset show that \systemname\ achieves state-of-the-art performance in static scene reconstruction, especially in challenging scenarios with large-scale, persistent occlusions.
\end{abstract}


\section{Introduction}
Reconstructing 3D scenes from monocular video remains a fundamental yet challenging problem in computer vision, particularly in real-world scenarios with dynamic foregrounds~\cite{mildenhall2021nerf, kerbl20233d}. The emergence of 3D Gaussian Splatting (3DGS)~\cite{kerbl20233d} has recently enabled high-fidelity 3D reconstruction with remarkable efficiency, leading to widespread adoption in applications such as robotics~\cite{sheng2025spatialsplat} and virtual reality~\cite{franke2025vr}.
In these settings, constructing a coherent and artifact-free static 3D environment is critical for downstream tasks, including perception~\cite{mallick2024taming, meng2023hybrid}, planning~\cite{lei2025gaussnav}, and simulation~\cite{feng2025flashgs}. However, real-world monocular videos often contain dynamic foreground objects that severely corrupt supervision, occlude large portions of the scene, and break temporal consistency across frames.
In this work, we aim to recover a clean and temporally consistent static background from monocular videos under persistent dynamic occlusions.

\begin{figure}[t]
    \centering
    \includegraphics[width=1.0\linewidth]{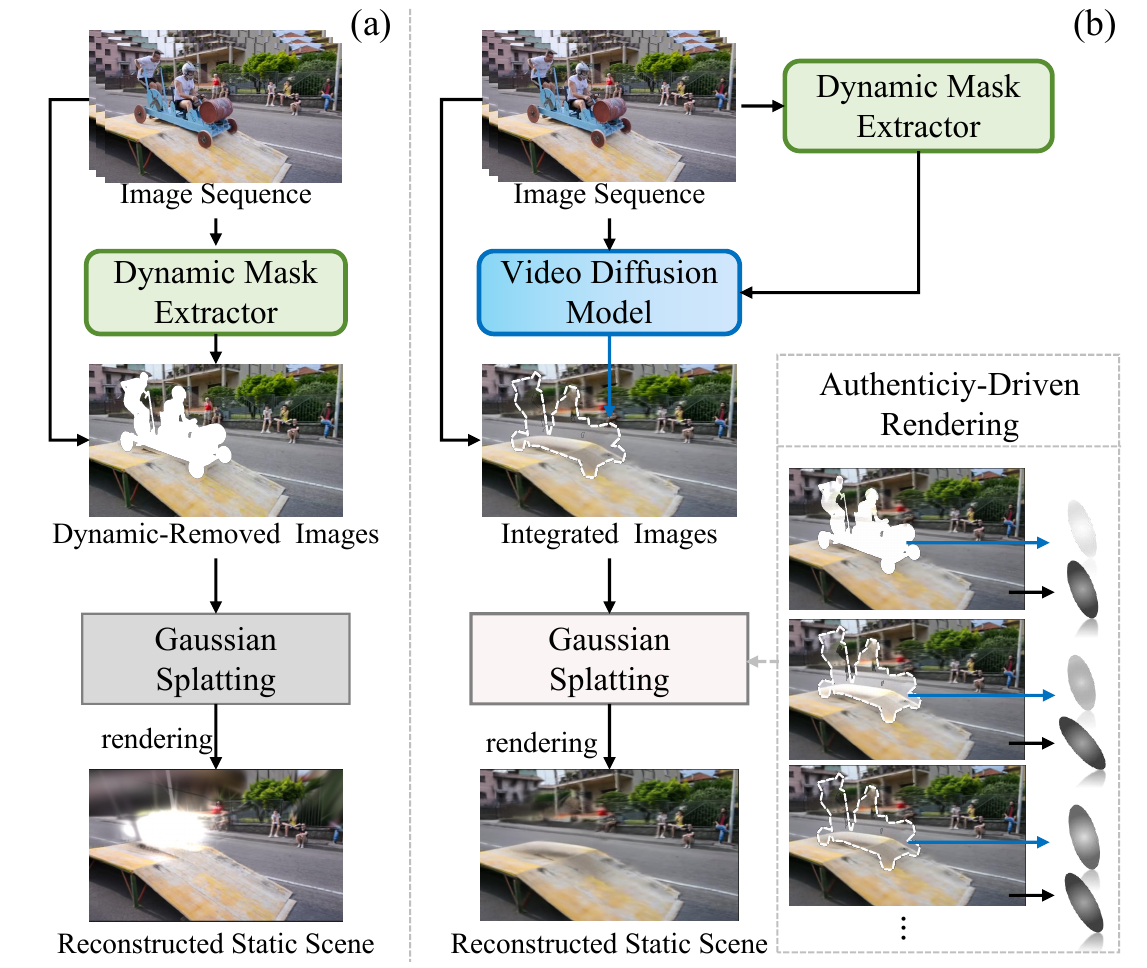} 
    \caption{\textbf{ Comparison between the previous pipeline (a) and our proposed \systemname~(b).} Previous methods supervise 3D Gaussian primitives solely based on background regions after dynamic object removal. In contrast, our \systemname~leverages a diffusion model to generate occluded content for auxiliary supervision and introduces \truth-driven rendering to balance real and generated information.}
    \label{fig:intro}
\end{figure}

As shown in Fig.~\ref{fig:intro}(a), existing approaches to static scene reconstruction typically remove dynamic objects and rely exclusively on the remaining background observations. Methods such as~\cite{wang2024gaussianeditor, liu2024infusion} directly manipulate Gaussian primitives, while others~\cite{kulhanek2024wildgaussians, ungermann2024robust} focus on suppressing transient distractors. However, these approaches are primarily designed for scenarios where dynamic objects occupy only a small portion of the scene or appear briefly. In more challenging real-world settings with large-scale and persistent dynamic occlusions, removing dynamic objects inevitably discards a substantial amount of valid supervision, resulting in incomplete or degraded reconstructions.
Alternatively, generative models offer a promising solution by hallucinating plausible content in occluded regions. Nevertheless, such generated content is inherently less reliable than real observations and may introduce inconsistencies. This leads to a fundamental dilemma: removing dynamic objects mitigates interference but sacrifices supervision, whereas introducing generative priors compensates for missing regions at the cost of incorporating unreliable pseudo-observations. Effectively balancing this trade-off remains an open challenge.

To address this challenge, we propose \systemname, a novel 3DGS-based framework for static scene reconstruction that jointly leverages real observations and generative priors. Our approach is built upon three key components as shown in Fig.~\ref{fig:intro}(b). First, we establish a reliable geometric foundation by initializing 3DGS from per-frame 3D estimates obtained via a 3D foundation model~\cite{wang2025vggt}, together with a structure-aware sampling strategy to filter noisy estimates and retain informative points. Second, to handle dynamic foregrounds in unconstrained videos, we introduce a motion-aware masking mechanism that combines the generalization capability of SAM~\cite{kirillov2023segment} with motion cues to accurately identify dynamic regions. Third, we incorporate diffusion-based video inpainting to generate auxiliary supervision for occluded regions.

Crucially, we observe that generated content is inherently less reliable than real observations. To explicitly model this distinction, we introduce a reliability-aware formulation that assigns each Gaussian primitive a learnable parameter, $\theta$, representing its likelihood of originating from real observations. This design enables the model to adaptively balance contributions from real and generated supervision, effectively leveraging generative priors while preserving reconstruction fidelity. As a result, our method resolves the trade-off between supervision completeness and reliability, leading to more accurate and temporally consistent static scene reconstruction under dynamic occlusions.

Furthermore, we note that existing dynamic scene datasets typically lack ground-truth static backgrounds, making it difficult to quantitatively evaluate reconstruction quality in occluded regions. To address this limitation, we construct a novel evaluation setup using a programmable robotic platform with a fixed camera trajectory to capture paired sequences: one containing dynamic objects and one with a clean static background, which serves as ground truth for evaluation.

Our key contributions include:
\begin{itemize}

\item We leverage multiple foundation models to construct a comprehensive pipeline for static 3D background reconstruction from monocular RGB videos. Our method achieves state-of-the-art performance on both the DAVIS dataset and our collected dataset.

\item We employ a diffusion model to inpaint dynamic regions, providing pseudo-ground-truth supervision for training. To effectively leverage both real background and generated content, we introduce a novel per-Gaussian primitive parameter that modulates opacity blending. This design preserves reconstruction fidelity while enhancing generalization by incorporating generative priors.

\item To quantitatively evaluate the ability of static scene reconstruction methods to recover occluded regions, we use a robot-mounted camera following a fixed trajectory to capture dynamic scenes and their corresponding static backgrounds, which serve as model inputs and ground-truth references, respectively.
\end{itemize}

\section{Related Work}
\subsection{Dynamic Scene Reconstruction}
In applications such as autonomous driving and embodied intelligence, dynamic content is inevitable in captured scenes~\cite{jiang2024gs,duan2026bugs}. Traditional 3DGS~\cite{kerbl20233d} struggles to reconstruct these dynamic regions directly, leading to the emergence of various dynamic scene reconstruction methods based on Gaussian splatting. To model 4D scenes, 4DGS~\cite{wu20244d} constructs Gaussian features from 4D neural volumes and uses a lightweight MLP to predict Gaussian deformations at new timestamps. For improved novel view synthesis, 4D Rotor gaussian~\cite{duan20244d} introduces anisotropic 4D XYZT Gaussians to efficiently represent spatiotemporal dynamics. Instead of modeling motion implicitly, Gaussian-Flow~\cite{Lin_2024_CVPR} employs a novel Dual-Domain Deformation Model to explicitly model the attribute deformations of each Gaussian point. In autonomous driving scenarios where multiple dynamic objects are present, methods like Street Gaussian~\cite{yan2024street}, OG-Gaussian~\cite{shen2025og}, and S3R-GS~\cite{zheng2025s3r} have proposed various strategies for decomposing scenes into static and dynamic components to enable more accurate reconstruction.

\subsection{Static Scene Reconstruction}
While directly reconstructing dynamic scenes is important, building a consistent and clean static 3D scene in environments with heavy dynamic interference remains a critical prerequisite for many downstream tasks. Several methods have been developed to reconstruct static backgrounds in the presence of dynamic interference. NeRF-based~\cite{mildenhall2021nerf} approaches such as RobustNeRF~\cite{liu2023robust} and Nerf on-the-go~\cite{ren2024nerf} have achieved promising results. With the emergence of 3DGS~\cite{kerbl20233d}, more efficient and low-cost reconstruction techniques have been proposed. HybridGS~\cite{lin2025hybridgs} introduces a hybrid framework that integrates multiview-consistent 3D Gaussians with independently modeled 2D Gaussians to separate static and dynamic elements. WildGaussians~\cite{kulhanek2024wildgaussians} combines powerful DINO~\cite{caron2021emerging} features with an appearance modeling module to robustly handle occlusions and appearance variations in 3DGS~\cite{kerbl20233d}. SpotlessSplat~\cite{sabour2025spotlesssplats} leverages Stable Diffusion priors and strong optimization techniques to suppress transient interference. To address long-term and large-scale dynamic objects, DAS3R~\cite{xu2024das3r} reconstructs the static background by incorporating predicted dynamic masks.

\section{Preliminary}

3D Gaussian Splatting is a scene reconstruction technique that represents the entire scene as a collection of Gaussian primitives. Each Gaussian primitive $G(\vec{x})$ is parameterized by its mean position $\vec{\mu}$, covariance matrix $\Sigma$, and opacity $\alpha$:
\begin{equation}
G(\vec{x})=\exp(-\frac{1}{2}(\vec{x}-\vec{\mu})^\top\Sigma^{-1}(\vec{x}-\vec{\mu}))
\end{equation}

To render the scene, the 3D Gaussian primitives are splatted onto the 2D image plane.
The contribution of each Gaussian primitive to the rendered image is determined by integrating its density along the viewing direction, weighted by its opacity $\alpha$ and color attribute $\vec{c}_i$:

\begin{equation}
\mathcal{C}_{\mathrm{u,v}}=\sum_{i=1}^N\vec{c}_i\alpha_i\prod_{j=1}^{i-1}(1-\alpha_j)
\end{equation}

$C_{u,v}$ denote the rendered color at pixel $(u,v)$, and $N$ is the number of all overlapping Gaussian primitives.
Differentiable rasterization blends their contributions into a continuous, trainable image.
All Gaussian parameters are optimized via gradient descent using the following loss:

\begin{equation}
\mathcal{L}_{\mathrm{loss}}=\mathcal{L}_{\mathrm{pixel}}+\lambda_{\mathrm{ssim}}\cdot\mathcal{L}_{\mathrm{ssim}}
\end{equation}

Here, $\mathcal{L}_{\text{pixel}}$ denotes the pixel-wise photometric loss, and $\mathcal{L}_{\text{SSIM}}$ captures the structural similarity loss.

\section{Method}

\begin{figure*}[thbp]
    \centering
    \includegraphics[width=1.0\linewidth]{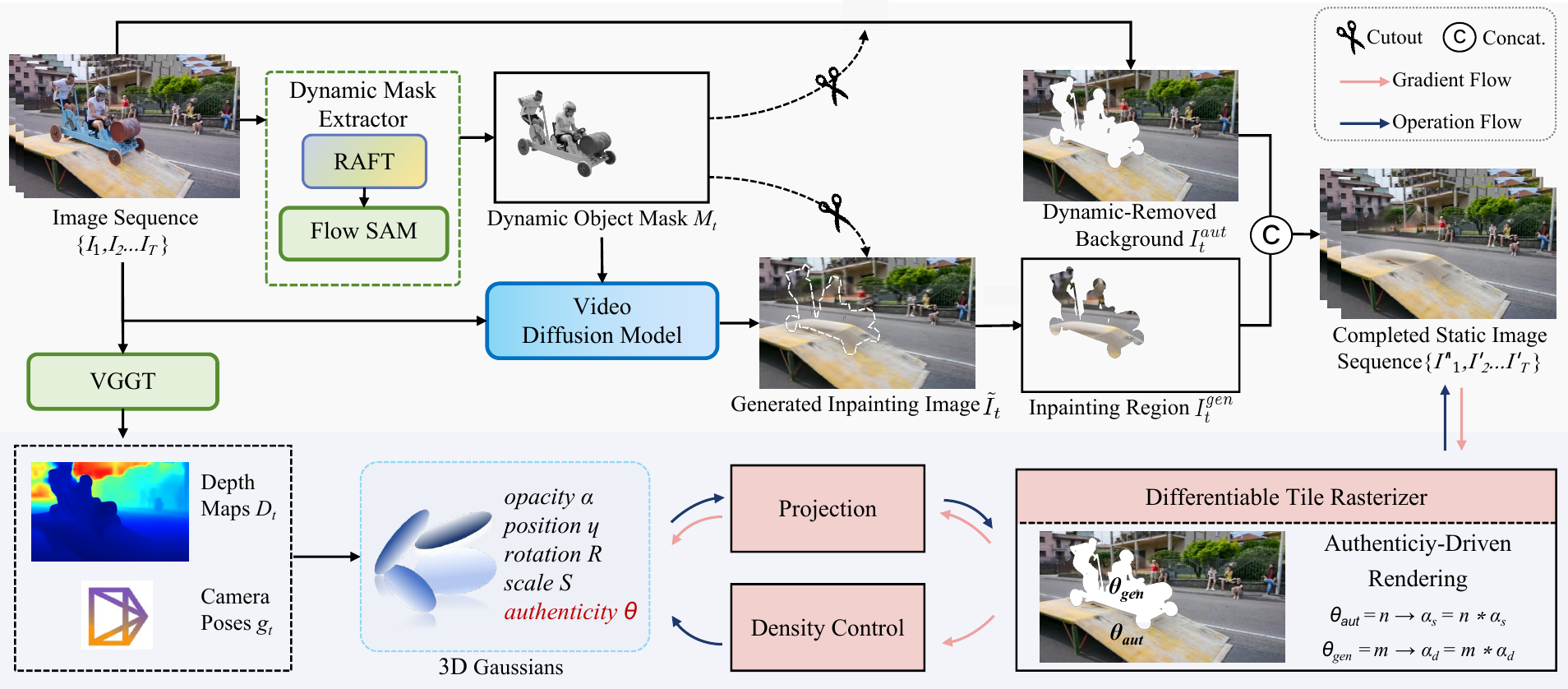} 
    \caption{\textbf{An overview of our \systemname\ pipeline.} We use VGGT~\cite{wang2025vggt} to obtain accurate camera poses and per-pixel 3D positions,  Then we employ a motion-aware SAM-based module to segment moving regions, and use a diffusion model to inpaint occlusions, providing pseudo-ground-truth supervision. In the opacity blending stage, the parameter $\theta$ is used to control the opacity of each Gaussian primitive, and the image space mask is applied to constrain the final loss.}
    \label{fig:framework}
\end{figure*}

\subsection{Problem Definition}
This work addresses the problem of static scene reconstruction from monocular video sequences captured in dynamic environments. Given an unlabeled video with freely moving objects, our goal is to recover the underlying static background geometry of the scene. To achieve this, we leverage 3D Gaussian Splatting as our scene representation, which facilitates high-quality rendering of the static components from novel viewpoints.

Let the input video be represented as a sequence of RGB image frames:

\[
\mathcal{I} = \{ I_i \mid i = 1, \dots, N \},
\]
where each frame satisfies $I_i \in \mathbb{R}^{H \times W \times 3}$ is captured from a dynamic video sequence with unknown camera intrinsics and poses. The sequence exhibits not only photometric and geometric variations induced by egocentric camera motion, but also significant occlusions and appearance changes caused by independently moving dynamic objects such as humans, vehicles, and animals. In this work, we assume no access to prior information about camera parameters (e.g., poses or focal lengths) or depth maps.

\subsection{Gaussian Initialization and Mask Extraction}
\noindent\textbf{Gaussian Initialization via VGGT. }Before performing 3D Gaussian-based scene reconstruction, it is necessary to provide a prior 3D point cloud for initializing the Gaussian primitives. Additionally, to project the 3D Gaussian primitives onto the corresponding camera views, camera parameters are required. Traditional methods rely on ground-truth calibrations or use Structure-from-Motion (SfM) approaches like COLMAP~\cite{schonberger2016structure}. In our work, we employ the end-to-end 3D foundation model VGGT~\cite{wang2025vggt} to estimate both the camera parameters and the initial point cloud used for training. This provides a practical initialization without requiring a separate SfM pipeline, and our method is not tied to this choice as shown in Sec.~\ref{sec:exp}.

We first capture a sequence of \( N \) RGB images \(\{I_i\}_{i=1}^N\) from the same scene. Then, we extract the required parameters from the VGGT model, using the transformer module:

\begin{equation}f\left((I_i)_{i=1}^N\right)=\left(R_i, \mathbf{t}_i, \mathbf{f}, D_i, D^\text{conf}_i\right)_{i=1}^N\end{equation}

In this way, the transformer can map each image \( I_i \) to its rotation $R_i$, translation $\mathbf{t}_i$, field of view $\mathbf{f}$, per-pixel depth map \( D_i \in \mathbb{R}^{H \times W} \) and a corresponding depth confidence map \( D^\text{conf}_i \). The depth map \( D_i \) associates each pixel location \( y \in \mathcal{I}(I_i) \) with a depth value \( D_i(y) \in \mathbb{R}^+ \). Similarly, the confidence map \( D^\text{conf}_i \) is also defined per pixel and indicates the model's confidence in its depth estimation. To better align with the camera model and projection geometry, we adopt depth-based back-projected points as priors for Gaussian primitives' initialization. We denote the image domain as \( \mathcal{I}(I_i) = \{1, \dots, H\} \times \{1, \dots, W\} \), representing the set of all pixel positions in the image. For any pixel \( y \in \mathcal{I}(I_i) \), its corresponding 3D point in space can be computed via a back-projection function:

\begin{equation}
\mathbf{X}_i(y) = R_i \cdot \Pi^{-1}(\mathbf{y}, D_i(\mathbf{y}), \mathbf{f}) + \mathbf{t}_i
\label{eq:backproj}
\end{equation}

Where $\Pi^{-1}$ represents the back projection function. By aggregating all the back-projected 3D points from images, we obtain a 3D point cloud \( \mathcal{X} \in \mathbb{R}^{N \times H \times W \times 3} \), where each point \( \mathbf{x} \in \mathbb{R}^3 \) corresponds to a pixel with coordinate \( (x, y, z) \) in the scene. To improve the reliability of these priors, we first filter the 3D points based on their corresponding depth confidence map \( D_i^{\text{conf}} \). While this improves point quality, the remaining set is still overly dense. Instead of random downsampling, we adopt a structure-aware sampling strategy that prioritizes points with high visual and geometric saliency. For each pixel \( y \), a sampling weight is defined as:

\begin{align}
S_i(y) =\; & 1\left[D_i^{\text{conf}}(y) > \tau \right] \cdot \Big( 
\lambda_1 \| \nabla I_i(y) \|_2 \nonumber \\
& \quad + \lambda_2 \| \nabla D_i(y) \|_2\Big)
\end{align}

where $\tau$ denotes the threshold for depth confidence, \( \nabla I_i(y) \) and \( \nabla D_i(y) \) are the image and depth gradients, respectively. We retain the top 10\% of points with highest \( S_i(y) \), resulting in a compact yet structurally informative set of initialization points for Gaussian primitives. This initialization stage reduces noise and redundancy in the input point cloud. 
After initialization, we follow the standard density control mechanism in 3D Gaussian Splatting~\cite{kerbl20233d}, 
which dynamically refines the Gaussian set via densification and pruning during training.

\noindent\textbf{Masks Generation. }
In our approach, we adopt Flow-SAM~\cite{xie2024moving} as the core mask generation module to obtain temporally consistent and motion-aware segmentation of dynamic regions. This module efficiently and accurately identifies and extracts binary masks of moving objects from input images. Flow-SAM integrates the strong generalization capability of the Segment Anything Model(SAM)~\cite{kirillov2023segment} with optical flow information, enabling it to focus on dynamic regions within the image. Specifically, given the input images \( \mathcal{I} \) and the corresponding optical flow \( \mathcal{F} \) (obtained using RAFT~\cite{teed2020raft}), Flow-SAM first utilizes its built-in trainable prompt generator to predict a set of prompts that effectively highlight moving objects based on \( \mathcal{F} \). These prompts are then fed into SAM to guide the segmentation. Finally, Flow-SAM outputs a set of separated binary masks \( \mathcal{M} = \{ M_i \}_{i=1}^N \):
\begin{equation}
\mathcal{M} = \text{Flow-SAM}(\mathcal{I}_{\text{RGB}}, \mathcal{F}) = \{M_i\}_{i=1}^N
\end{equation}
Here, \( N \) denotes the number of detected moving objects, and each \( M_i \) represents a pixel-wise precise segmentation mask corresponding to an individual moving object.

\begin{figure*}[thbp]
    \centering
    \includegraphics[width=1.0\linewidth]{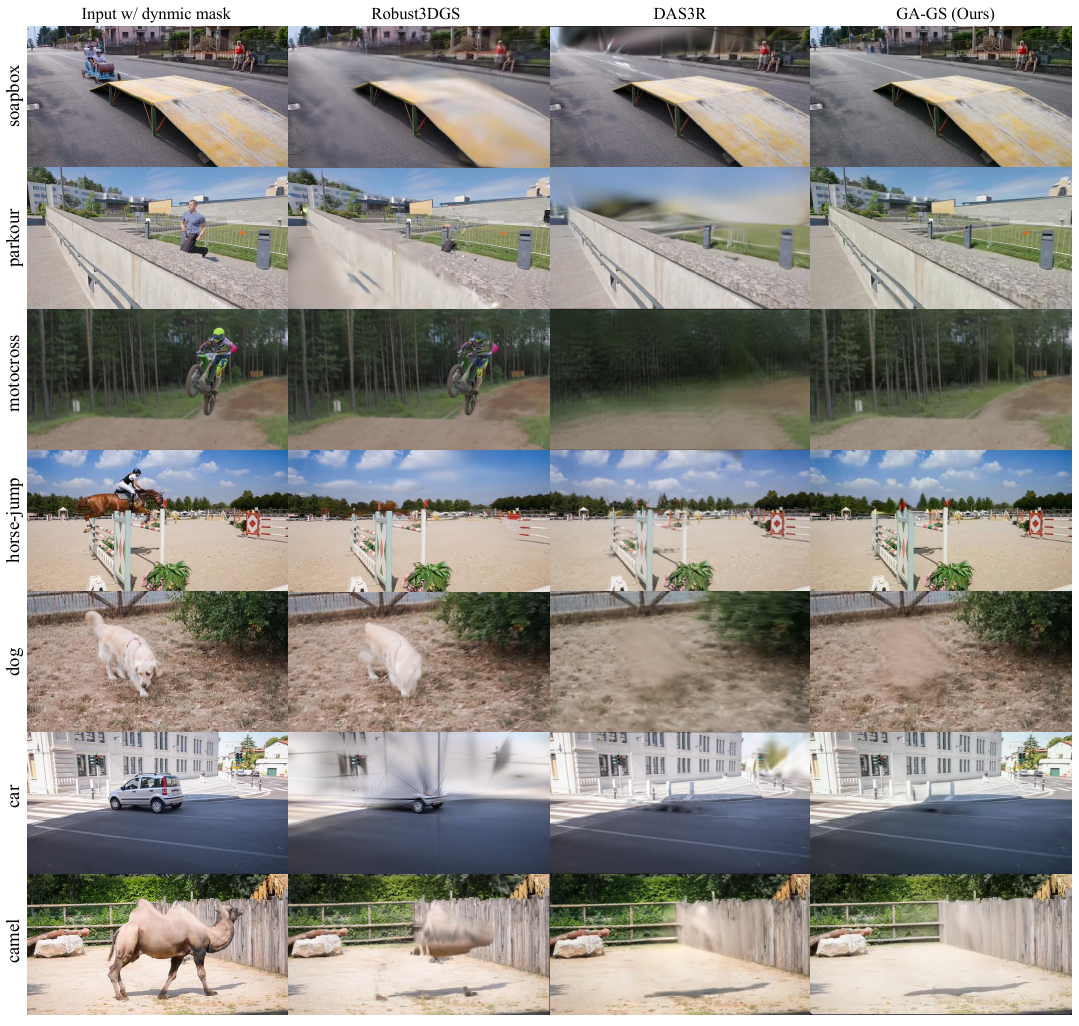} 
    \caption{\textbf{Visualization on the DAVIS dataset. }Since ground truth for the static background is unavailable, the first row only shows the input containing dynamic objects. Compared to the baselines, our method achieves better visual results in both background reconstruction and occlusion region recovery.}  
    \label{fig:davis_exp}
\end{figure*}

\subsection{Pseudo-Ground Truth Generation}
To provide a clean reference for downstream static background modeling, we utilize DiffuEraser~\cite{li2025diffueraser}, a diffusion-based video inpainting model, to remove dynamic foreground objects from the input video in a mask-guided manner. DiffuEraser is capable of generating structurally coherent and texture-rich content even in large occluded regions, while maintaining good temporal consistency. Given an input video \( \mathcal{I} = \{I_i\}_{i=1}^N \) and its corresponding binary masks \( \mathcal{M} = \{M_i\}_{i=1}^N \), where \( M_i(y) = 1 \) indicates that pixel \( y \) in frame \( i \) is occluded, each frame-mask pair \( (I_i, M_i) \) is fed into DiffuEraser to obtain the inpainted result:

\begin{equation}
\tilde{I}_i = \text{DiffuEraser}(I_i, M_i)
\end{equation}

To ensure that the supervision data remains as close as possible to the original input, we extract the inpainted regions from the DiffuEraser-generated video using the corresponding binary masks, and blend them back into the original image sequence \( \mathcal{I} = \{I_i\}_{i=1}^N \). This produces a new supervision image sequence \( \mathcal{I}' = \{I_i'\}_{i=1}^N \), where each supervised frame \( I_i' \) is defined as:

\begin{equation}
I_i'(y) = M_i(y) \cdot \tilde{I}_i(y) + (1 - M_i(y)) \cdot I_i(y)
\label{eq:blend}
\end{equation}

Here, \( \tilde{I}_i \) denotes the inpainted frame produced by DiffuEraser, \( M_i \in \{0,1\}^{H \times W} \) is the binary mask indicating the foreground region in frame \( i \), and \( y \in \mathcal{I}(I_i) \) is a pixel location within the image domain. This operation ensures that only the masked pixels are replaced by the inpainted result.

\subsection{Authenticity-aware Opacity Blending}

Once the initial 3D point cloud and camera parameters are obtained from VGGT, we proceed to initialize 3D Gaussian primitives. Each selected 3D point serves as a position prior for one Gaussian primitive.

However, it is important to note that our supervision images \( \mathcal{I}' = \{I_i'\}_{i=1}^N \) are constructed by blending the inpainted regions with the original frames. Consequently, not all pixels in \( \mathcal{I}' \) originate from the true scene content. To explicitly represent this distinction, we assign each initialized Gaussian primitive an additional scalar attribute \( \theta \in [0, 1] \), referred to as the \truth~indicator.
This attribute encodes likelihood that the Gaussian primitive originates from an unaltered (real) region of original video. Specifically, we initialize \( \theta = 0.9 \) if corresponding pixel in \( I_i' \) is from original image, and \( \theta = 0.1 \) if it is from the inpainted region. The initial \truth~value \( \theta_i(y) \) is inherited by the Gaussian primitive initialized at location \( \mathbf{X}_i(y) \). Importantly, this scalar \( \theta \in [0, 1] \) is treated as learnable parameter and is jointly optimized with Gaussian position, opacity, scale, rotation, and color parameters during training. This allows the model to adaptively refine the contribution of each Gaussian primitive, serving both as prior and a modulation factor for rendering and loss computation.

During the final image synthesis stage, we modulate the effective opacity of each Gaussian primitive by its associated \( \theta \in [0, 1] \). This mechanism ensures that Gaussian primitives originating from real regions contribute more dominantly to rendering, while those from inpainted areas serve as auxiliary support. Specifically, for a given Gaussian primitive with original opacity \( \alpha \), the adjusted opacity \( \alpha' \) used during blending becomes: $\alpha' = \theta \cdot \alpha$, \( \mathbf{c}_i \) represents the color of the \( i \)-th Gaussian primitive, the final rendered color is computed as:

\begin{equation}
\begin{split}
\mathbf{C}_{\text{final}} &= \sum_{i=1}^{n} \alpha_i' \cdot \mathbf{c}_i \cdot \prod_{j=1}^{i-1} (1 - \alpha_j') , \\
\end{split}
\end{equation}

During the training phase, each supervision frame \( I_i' \) is used for loss computation, regardless of whether the pixel belongs to the real or inpainted region. This ensures that the model learns to reconstruct the entire scene comprehensively. To control the contribution of different regions in the loss, we compute a weighted L1 loss between the rendered image \( \hat{I}_i \) and the supervision image \( I_i' \). Composite weight map \( w_i \) is derived from the binary mask \( M_i \), which assigns higher weight to real regions and lower weight to inpainted ones.
The pixel-wise L1 loss is defined as:

\begin{equation}
\mathcal{L}_{\text{L1}} = \sum_{y \in \mathcal{I}(I_i)} w_i(y) \cdot \left| \hat{I}_i(y) - I_i'(y) \right|
\end{equation}

where the composite weight map \( w_i(y) \) is defined as:

\begin{equation}
w_i(y) = 1.0 - (1-w) \cdot M_i(y)
\end{equation}

Here, \( M_i(y) = 0 \) indicates a real (unmodified) pixel and \( M_i(y) = 1 \) indicates a generated (inpainted) pixel, $w \in [0,1]$ controls the relative contribution of generated regions. As a result, real pixels are given full weight (1.0), while generated pixels receive a lower weight ($w$). This encourages the model to prioritize the reconstruction of authentic scene content, while still leveraging information from inpainted regions for improved generalization.

To ensure numerical stability during training, the SSIM loss value \( \mathcal{L}_{\text{SSIM}} \) is clamped to the valid range \([0, 1]\) before participating in the final loss computation. This prevents potential negative values due to floating-point imprecision. The total reconstruction loss combines the weighted L1 loss and the clamped SSIM loss as follows:

\begin{equation}
\mathcal{L}_{\text{total}} = \lambda_{\text{L1}} \cdot \mathcal{L}_{\text{L1}} + \lambda_{\text{SSIM}} \cdot \text{Clamp}(\mathcal{L}_{\text{SSIM}}, 0, 1)
\end{equation}

\section{Experiment}
\label{sec:exp}
In following sections, we analyze performance of our method. We compare it with existing static scene reconstruction approaches on two datasets and conduct a series of ablation studies to evaluate the effectiveness of each component.

\begin{table*}[th]
    \centering
    \caption{\textbf{Quantitative results on DAVIS dataset. }Eight scenes from dataset are selected for evaluation. PSNR is computed over background regions, as defined by the ground-truth dynamic masks. \textbf{Bold} values indicate best method in average.}
    \renewcommand{\arraystretch}{1.8}
\normalsize
    \begin{tabular}{
    >{\centering\arraybackslash}m{3.1cm}|
    >{\centering\arraybackslash}m{1.3cm}
    >{\centering\arraybackslash}m{1.0cm}
    >{\centering\arraybackslash}m{1.0cm}
    >{\centering\arraybackslash}m{1.0cm}
    >{\centering\arraybackslash}m{1.4cm}
    >{\centering\arraybackslash}m{1.4cm}
    >{\centering\arraybackslash}m{1.0cm}
    >{\centering\arraybackslash}m{1.1cm}|
    >{\centering\arraybackslash}m{1.1cm}
    }
        \hline
        
        \rowcolor[gray]{.92}
           & Blackswan & Camel & Car-shadow & Dog & Horse-jump & Motocross-jump & Parkour & Soapbox & \textbf{Average} \\
           \hline
        \hline
         WildGaussians \small{~\cite{kulhanek2024wildgaussians}} & 18.95 & 19.19 & 21.45 & 19.74 & 18.79  & 7.91  & 18.89 & 20.55 & 18.18  \\
         
          Robust3DGS \small{~\cite{ungermann2024robust}} & 19.58 & 21.31 & 29.31 & 22.48 & 20.87 & 13.83 & 21.29 & 22.55 & 21.40 \\ 
          
          DAS3R \small{~\cite{xu2024das3r}} & 23.90 & 27.27 & 29.13 & 28.63 & 25.09 & 17.09 & 28.09 & 26.41 & 25.70 \\
          
          \textbf{\systemname~(Ours)} & 25.75 & 27.69 & 31.95 & 29.15 & 25.67 & 20.42 & 28.74 & 29.72 & \textbf{27.39} \\
          \bottomrule
    \end{tabular} 
    \label{tab:davis}
\end{table*}

\begin{figure}
    \centering
    \includegraphics[width=0.8\linewidth]{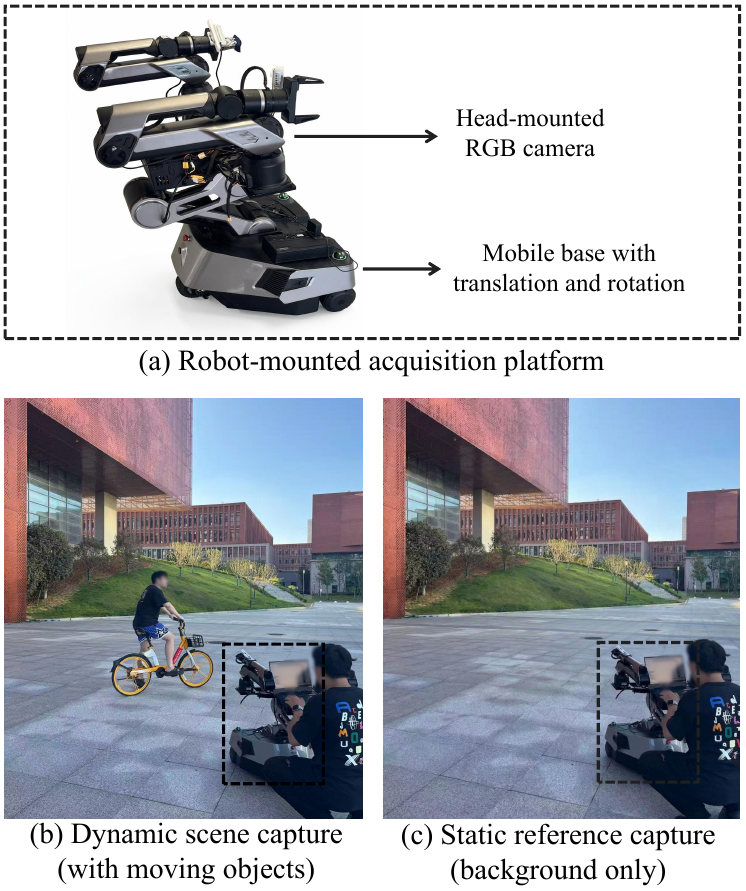}
    \caption{\textbf{Data acquisition process of the Trajectory-Match dataset.} 
    (a) We employ a robot-mounted camera platform to ensure precise and repeatable camera trajectories. 
    (b) For each scene, we first capture a dynamic sequence containing moving objects such as pedestrians or vehicles. 
    (c) We then record a corresponding static sequence along the same trajectory after    removing dynamic elements, which serves as ground-truth background. 
    This paired acquisition setup enables direct quantitative evaluation of reconstruction performance in occluded regions.}
    \label{fig:record}
\end{figure}

\noindent\textbf{Dataset.} We evaluate our method on the DAVIS~\cite{Perazzi2016} and our \ourdata.
DAVIS is widely used for video object segmentation tasks. We select 8 challenging sequences characterized by long-term and large-area dynamic object presence. While DAVIS provides accurate ground-truth masks for dynamic objects, it does not contain ground-truth static backgrounds, making it unsuitable for evaluating reconstruction quality in occluded regions.
To enable quantitative evaluation of occluded region reconstruction, we construct the Trajectory-Match dataset using a robot-mounted acquisition platform. Specifically, we employ a robot equipped with a head-mounted RGB camera and a mobile base supporting both translation and rotation, which allows precise and repeatable camera trajectories. As illustrated in Fig.~\ref{fig:record}, for each scene, we first capture a dynamic sequence containing moving objects such as pedestrians, bicycles, and vehicles, and then record a corresponding static sequence along the same trajectory after removing all dynamic elements. 
Our dataset covers indoor, outdoor, and street scenarios. This paired acquisition setup ensures accurate spatial alignment between dynamic and static sequences, enabling direct pixel-wise comparison between reconstructed results and ground-truth static backgrounds, particularly in regions that are occluded in the dynamic input.

\noindent\textbf{Baselines.} We compare our method with WildGaussians~\cite{kulhanek2024wildgaussians}, Robust3DGaussians~\cite{ungermann2024robust}, and DAS3R~\cite{xu2024das3r}. WildGaussian and RobustGaussian are designed primarily for distractor-free static scene reconstruction, DAS3R~\cite{xu2024das3r} is capable of handling scenes with long-term dynamic object presence. 

\begin{figure}[thbp]
    \centering
    \includegraphics[width=1.0\linewidth]{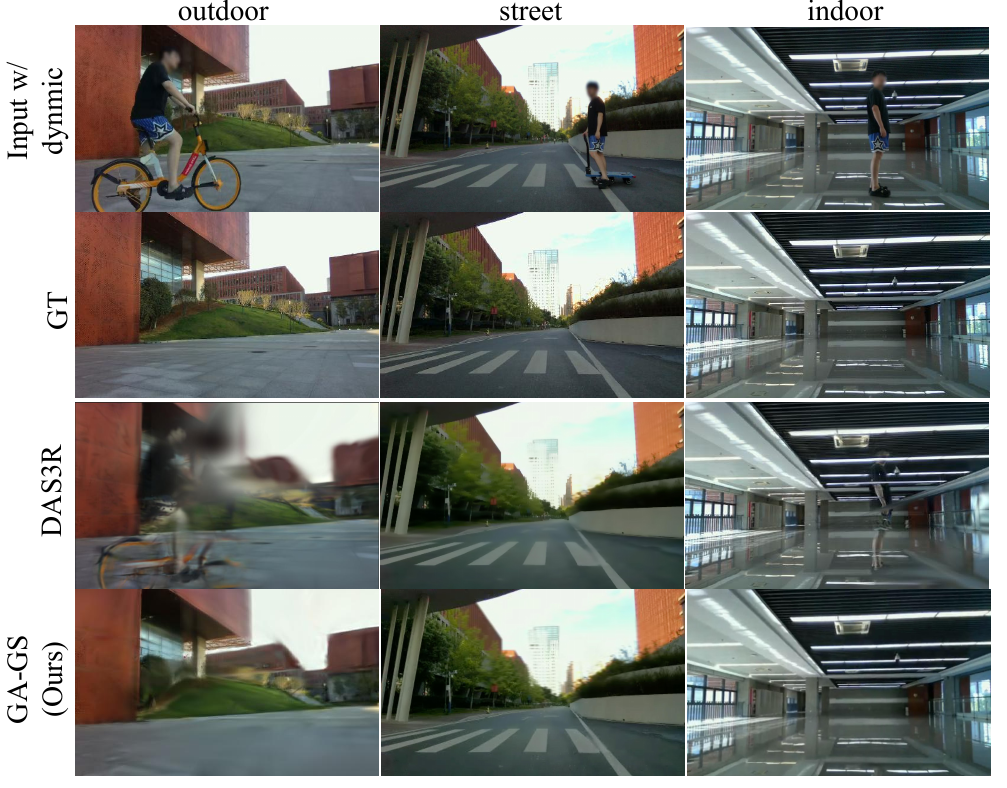} 
    \caption{\textbf{Visualizations on \ourdata. }The second row presents the ground truth of recorded static scene, serving as a reference for comparing the reconstructions of \systemname \ and the baseline in the third and fourth rows. }
    \label{fig:real_world}
    \vspace{-2mm}
\end{figure}

\noindent\textbf{Metrics.} We use three commonly adopted metrics for image reconstruction: 
PSNR (Peak Signal-to-Noise Ratio) for pixel-level accuracy, 
SSIM (Structural Similarity Index) for structural consistency, 
and LPIPS (Learned Perceptual Image Patch Similarity) for perceptual similarity based on deep features.

\noindent\textbf{Implementation details. }All experiments are conducted on a single NVIDIA A100 GPU. For camera parameter estimation, we employ the pretrained VGGT-1B model. During the Gaussian training stage, the optimization is performed for 4000 iterations. The learning rates are set to 0.00016 for position, 0.005 for scaling, and 0.05 for opacity. The degree of spherical harmonics is set to the default value of 3. All external modules, including VGGT, RAFT, Flow-SAM, and DiffuEraser, are used as pretrained models and remain frozen during training.

\subsection{Static Scene Reconstruction}
We conduct experiments on eight sequences from the DAVIS~\cite{Perazzi2016} dataset and three scenes from our \ourdata. Figures~\ref{fig:davis_exp} and Table~\ref{tab:davis} present qualitative and quantitative results on the DAVIS dataset~\cite{Perazzi2016}, respectively. Since ground-truth images of the static scene are not available in this dataset, we use the ground-truth dynamic masks to exclude dynamic objects when computing quantitative results, evaluating only the reconstruction quality of the background. However, due to the irregular shapes of the background, SSIM and LPIPS are less reliable in such regions. Therefore, we report only PSNR for these area. Our method outperforms all baselines in terms of PSNR for background reconstruction. Qualitatively, it achieves significantly better recovery in occluded regions, such as those in the car-shadow and camel scenes. In more challenging cases involving large occlusions and complex motion(e.g., the motocross scene), our approach still maintains superior reconstruction quality compared to existing methods.

\begin{table*}[th]
    \centering
    \caption{\textbf{Comparison of static scene reconstruction performance on \ourdata.} Results are provided for full scene static reconstruction (indoor, outdoor, street, average) and dynamic occluded region reconstruction (indoor-occluded, outdoor-occluded, street-occluded, average-occluded). \textbf{Bold} values highlight the best-performing method in average.}
    \renewcommand{\arraystretch}{1.2}
    \setlength{\tabcolsep}{4pt}
    \begin{tabular}{>{\centering\arraybackslash}m{3.2cm}|>{\centering\arraybackslash}m{1.2cm}|>{\centering\arraybackslash}m{1.2cm}|>{\centering\arraybackslash}m{1.2cm}|>{\centering\arraybackslash}m{1.2cm}|c|>{\centering\arraybackslash}m{1.5cm}|>{\centering\arraybackslash}m{1.5cm}|>{\centering\arraybackslash}m{1.5cm}|>{\centering\arraybackslash}m{1.5cm}}
    
    \hline
        \rowcolor[gray]{.92}
        
         & Metrics & Indoor & Outdoor & Street & Average & Indoor occluded & Outdoor occluded & Street occluded & Average occluded \\
         \hline
         \hline
         
         &PSNR$\uparrow$ & 19.47 & 20.10 & 21.55 & 20.37 & 15.82 & 15.07 & 15.29 & 15.39\\
         WildGaussians  & SSIM$\uparrow$ & 0.66 & 0.65 & 0.77 & 0.69 & - & - & - & -  \\
         \small{~\cite{kulhanek2024wildgaussians}} & LPIPS$\downarrow$  & 0.27 & 0.42 & 0.37 & 0.35 & - & - & - & - \\
         \hline
         &PSNR$\uparrow$ & 19.45 & 18.89 & 20.61 & 19.65 & 16.11 & 15.73 & 14.98 & 15.61\\
        Robust3DGS &SSIM$\uparrow$ & 0.61 & 0.58 & 0.70 & 0.63 & - & - & - & - \\
        \small{~\cite{ungermann2024robust}} &LPIPS$\downarrow$ & 0.26 & 0.31 & 0.24  & 0.27 & - & - & - & - \\
        \hline
         & PSNR$\uparrow$ & 21.45 & 20.07 & 22.31 & 21.28 & 16.40 & 14.22 & 20.18 & 16.93 \\
        DAS3R & SSIM$\uparrow$ & 0.74 & 0.68 & 0.77 & 0.73 & - & - &- & -\\
        \small{~\cite{xu2024das3r}} & LPIPS$\downarrow$ & 0.18 & 0.29 & 0.17 & 0.21 & - & - & - & - \\
        \hline
        & PSNR$\uparrow$& 25.34 & 24.67 & 25.62 & \textbf{25.21} & 24.91 & 24.36 & 25.13 & \textbf{24.80} \\
        \textbf{\systemname~(Ours)}  &SSIM$\uparrow$& 0.86 & 0.84 & 0.87 & \textbf{0.86} & - & - & - & - \\
        &LPIPS$\downarrow$& 0.10 & 0.14 & 0.09 & \textbf{0.11} & - & - & - & - \\
        \bottomrule
    \end{tabular}
    \label{tab:real-world}
    \vspace{-4mm}
\end{table*}

Table~\ref{tab:real-world} presents the comparison results for static scene reconstruction on \ourdata. For this dataset, we compute three metrics between rendered outputs and the ground-truth static images. Additionally, we separately evaluate the occluded regions. Again, due to irregularity of the masks, we report only PSNR for occluded areas (results with the "occluded" suffix in the table). This targeted evaluation provides a clearer indication of effectiveness of our method in mitigating the impact of dynamic regions. Our approach outperforms all baseline methods in terms of PSNR, SSIM, and LPIPS metrics, particularly demonstrating significant improvements in PSNR within the occluded regions. Figure~\ref{fig:real_world} shows qualitative results on \ourdata. It is evident that our method achieves higher reconstruction quality in complex scenes, especially in indoor and outdoor environments with dynamic object motion. Compared to baseline methods, our approach exhibits notably better recovery of occluded regions, with almost no residual artifacts remaining.

\subsection{Fairness and Pose Source Analysis}
To examine whether the performance gain is influenced by the choice of pose estimation and initialization source, we compare COLMAP and VGGT for both DAS3R and our method in Table~\ref{tab:pose_ablation}. While COLMAP consistently achieves slightly better performance due to its more accurate multi-view optimization, our method still outperforms DAS3R under both pose sources, indicating that the improvement is not tied to a specific initialization pipeline. Notably, the performance gap between COLMAP and VGGT is smaller in our method, demonstrating stronger robustness to pose and depth noise. We adopt VGGT in our default pipeline because it provides a fast, end-to-end solution for jointly estimating camera parameters and dense geometry without requiring multi-view optimization, making it more practical for real-world or online scenarios. These results confirm that our gains primarily stem from the proposed generation-assisted design rather than the choice of pose estimation.

\subsection{Ablations}
The significance of individual components is examined through targeted ablation studies. Emphasis is placed on evaluating the influence of \truth~parameter and the diffusion-based generation module, both essential to handling occlusions and ensuring high-fidelity reconstruction.

\noindent\textbf{Effect of the Authenticity Parameter.}  To evaluate the contribution of this new parameter, we conduct an ablation study on \ourdata~ by removing it from the model, while keeping all other settings unchanged. As shown in Table~\ref{tab:ablation}, this modification leads to a notable PSNR drop in occluded regions, with performance decreasing from 24.80 to 23.85, indicating a significant degradation. This parameter is designed to guide the model in assigning higher confidence to visible content, especially under partial occlusions. Without it, the generated content would be assigned the same weight, causing the reconstruction to overly rely on the generated content. Table~\ref{tab:ablation} also includes a baseline that removes both generation and authenticity modeling. The resulting performance drop confirms that relying solely on masked observations is insufficient under persistent occlusion, and that the proposed generation and authenticity mechanisms are complementary.

\begin{table}[]
    \centering
    \caption{\textbf{Quantitative results of ablation experiments on \truth\ parameter and generation module.} 
'Occl.' refers to occluded regions, $\theta$ denotes \truth\ parameter. 
The column ``w/o both'' corresponds to removing both generation $\theta$, i.e., using masked observations for training.}
    \begin{tabular}{l|
    >{\centering\arraybackslash}m{1.1cm}
    >{\centering\arraybackslash}m{1.2cm}
    >{\centering\arraybackslash}m{1.4cm}
    >{\centering\arraybackslash}m{1.2cm} }

         \hline
         \rowcolor[gray]{.92}
         & w/o both & \systemname\ w/o $\theta$ & \systemname\ w/o generation & \systemname \\
         \hline
         \hline
        Indoor & 24.32 & 24.68 & 24.97 & 25.34 \\
        Outdoor & 23.72 & 24.07 & 24.13 & 24.67 \\
        Street & 24.18 & 24.53 & 25.19 & 25.62 \\
        Average  & 24.07 & 24.43 & 24.76 & \textbf{25.21}\\
        \hline
        Indoor-occl.  & 23.52 & 23.89 & 24.43 & 24.91 \\
        Outdoor-occl. & 23.05 & 23.45 & 23.90 & 24.36 \\
        Street-occl. & 23.82 & 24.22 & 24.65 & 25.13 \\
        Average-occl. & 23.46 & 23.85 & 24.33 & \textbf{24.80} \\
        \hline
    \end{tabular} 
    \label{tab:ablation}
\end{table}

\begin{table}[t]
\centering
\caption{\textbf{Impact of pose and initialization source.} 
We compare COLMAP and VGGT for both our method and DAS3R.}
\label{tab:pose_ablation}
\resizebox{\linewidth}{!}{
\begin{tabular}{l|ccccc}
\hline
\rowcolor[gray]{.92}
Method & Pose & Indoor & Outdoor & Street & Avg \\
\hline
\multirow{2}{*}{DAS3R} & COLMAP & 21.45 & 20.07  & 22.31 & 21.28 \\
 & VGGT & 20.78 & 19.31 & 21.54 & 20.54 \\
\hline
\multirow{2}{*}{Ours} & COLMAP & 25.61 & 25.03 & 25.88 & 25.51 \\
 & VGGT & 25.34 & 24.67 & 25.62 & 25.21 \\
\hline
\end{tabular}}
\end{table}

\begin{table}[t]
\centering
\caption{\textbf{Sensitivity to authenticity initialization.} 
We vary the initial authenticity values assigned to Gaussian primitives from real and generated regions, respectively.}
\label{tab:param_ablation}
\begin{tabular}{c|cccc}
\hline
\rowcolor[gray]{.92}
Setting & Full PSNR & Full SSIM & Full LPIPS & Occl. PSNR \\
\hline
(0.5, 0.5)   & 24.46 & 0.84 & 0.13 & 23.92 \\
(0.7, 0.3)   & 24.81 & 0.85 & 0.12 & 24.29 \\
(0.8, 0.2)   & 25.03 & 0.85 & 0.11 & 24.56 \\
(0.9, 0.1)   & 25.21 & 0.86 & 0.11 & 24.80 \\
(0.95, 0.05) & 25.08 & 0.85 & 0.12 & 24.48 \\
\hline
\end{tabular}
\end{table}

\noindent\textbf{Effect of the Diffusion-Based Generation.} We further evaluate the importance of diffusion-based generation module, which synthesizes plausible static background content in regions occluded by dynamic objects. As shown in Table~\ref{tab:ablation}, removing this module leads to a clear drop in reconstruction quality — the average PSNR decreases from 25.21 to 24.76, with a similarly noticeable decline in performance within occluded regions. These results highlight the importance of the module in handling areas with severe occlusion, where direct visual information is missing. By leveraging generative models to infer missing content, the module plays a crucial role in maintaining reconstruction fidelity.

\subsection{Sensitivity to Authenticity Initialization}
We further study sensitivity of the proposed authenticity prior by varying initialization values assigned to Gaussian primitives from real and generated regions. As shown in Table~\ref{tab:param_ablation}, using identical initialization for both sources, i.e., $(0.5,0.5)$, yields weakest performance, indicating source-aware initialization is necessary. Increasing separation between real and generated priors steadily improves reconstruction quality, and the default setting $(0.9,0.1)$ achieves best overall trade-off. When gap becomes overly large, e.g., $(0.95,0.05)$, performance slightly drops, especially in occluded regions, suggesting excessively suppressing generated regions weakens their value as auxiliary supervision.

\section{Conclusion}
In this work, we introduce \systemname, a novel 3D Gaussian Splatting-based framework for static scene reconstruction from monocular videos containing dynamic foreground objects. \systemname\ leverages motion-aware segmentation to accurately identify dynamic regions, and employs diffusion-based inpainting to synthesize occluded background content, providing effective pseudo-ground-truth supervision. A learnable \truth\ scalar is incorporated to dynamically balance contributions from real and generated content via opacity-aware rendering, enhancing reconstruction fidelity. To support comprehensive evaluation, we collected a real-world dataset consisting of paired dynamic and clean background videos. Extensive experiments conducted on both the DAVIS benchmark and our dataset demonstrate that \systemname\ consistently achieves state-of-the-art performance, particularly excelling in challenging scenarios with heavy and persistent occlusions.

\clearpage

\bibliographystyle{IEEEtran}
\bibliography{reference}


\end{document}